\documentclass{article}
\date{}
\title{A Scheme for Dynamic Risk-Sensitive Sequential Decision Making}	
\author{Shuai Ma$ ^{*} $ \and Jia Yuan Yu\thanks{S. Ma and J. Y. Yu are with the Concordia Institute of Information System Engineering, Concordia University, Canada. Emails: {\tt\small m\_shua@encs.concordia.ca}, {\tt\small jiayuan.yu@concordia.ca}} \and Ahmet Satir\thanks{A. Satir is with the John Molson School of Business, Concordia University, Canada. Email: {\tt\small ahmet.satir@concordia.ca}}%% <-this % stops a space
}

\usepackage{amsfonts,amsmath,xcolor,graphicx,dsfont,url,comment,soul}
\usepackage{algorithmicx, algpseudocode}  %  write algorithms in pseudo code in a uniform style
\usepackage{algorithm} 

\newtheorem{theorem}{Theorem}

\newtheorem{definition}{Definition}
\newtheorem{problem}{Problem}

\DeclareMathOperator{\supp}{supp}

\begin{document}
	
	\maketitle
    
%	\thispagestyle{empty}
%	\pagestyle{empty}

%%%%%%%%%%%%%%%%%%%%%%%%%%%%%%%%%%%%%%%%%%%%%%%%%%%%%%%%%%%%%%%%%%%%%%%%%%%%%%%%
\begin{abstract}
	
We present a scheme for sequential decision making with a risk-sensitive objective and constraints in a dynamic environment. % (SDM)
A neural network is trained as an approximator of the mapping from parameter space to space of risk and policy with risk-sensitive constraints. %some specified risk measure(s).
For a given risk-sensitive problem, in which the objective and constraints are, or can be estimated by, functions of the mean and variance of return, we generate a synthetic dataset as training data.
Parameters defining a targeted process might be dynamic, i.e., they might vary over time, so we sample them within specified intervals to deal with these dynamics. %vary over time
We show that: i). Most risk measures can be estimated using return variance; ii). By virtue of the state-augmentation transformation, practical problems modeled by Markov decision processes with stochastic rewards can be solved in a risk-sensitive scenario; and iii). The proposed scheme is validated by a numerical experiment. 
	
\end{abstract}

%%%%%%%%%%%%%%%%%%%%%%%%%%%%%%%%%%%%%%%%%%%%%%%%%%%%%%%%%%%%%%%%%%%%%%%%%%%%%%%%
\section{INTRODUCTION}
%\color{blue}
Cities are constantly involved in complex changing processes. 
City logistics is characterized by multiple stakeholders with different objectives and constraints.
Sustainability plays an increasingly significant role in planning and management within organizations and across supply chains~\cite{Grzybowska2014}. %Grzybowska2014
The sustainable city logistics studies express reflections and developments on the economic and the environmental problems on supply chain management, but only few of them take risk into consideration.
Since many sustainable city logistics problems, such as planning, routing and scheduling, can be modeled as sequential decision making (SDM) problems~\cite{Franceschetti2015}, %Franceschetti2015
in this paper, we propose a scheme composed of reinforcement learning (RL) methods for risk-sensitive SDMs with constraints in dynamic environments.

\color{black}
Risk-sensitive objectives have been drawing more attention in many practical problems, especially in which the small probability events have serious consequences.
Since various goals are targeted in practice, one solution is to optimize one of them as objective and take others as constraints.
Furthermore,  the environment might vary over time, so we sample the variables involved in the decision-making within specified intervals to deal with these dynamics.
In this paper, we consider the SDM problems with three concerns: risk, constraint, and dynamic environment.
In order to solve this problem, we propose a scheme for generating a synthetic dataset and training an approximator (here we use neural network as an example). %for solving such SDM problems with risk-sensitive objectives and constraints.
%model the environment for each the risk evaluation in constrained SDM problems with stochastic rewards in dynamic environments,
%The motivation for such schemes arises from complex problems for which exact analytic modeling and analysis may be excessively complicated or prohibitively expensive
%We study risk in constrained Markov decision processes (CMDPs) with stochastic reward functions.
The motivation for the proposed scheme arises from practical problems in which exact analytic risk analysis may be excessively complicated or prohibitively expensive, especially in a dynamic scenario.
Though historical data is usually used for neural network training, in many practical situations, the recorded decisions are not optimal.
Furthermore, in risk-sensitive cases, the criteria based on which the decisions were made are not clarified. %in the data records
Therefore, %the historical data cannot be used for NN training in these cases.
we generate a synthetic dataset for problems with specified risk-sensitive objective and constraints.
For a given parameter set, we construct a Markov decision process (MDP), calculate the return variance for any deterministic policy, and then evaluate (or estimate) the specified risk measures and record the optimal policy. %or the risk requirement
%By doing so, we generate a synthetic dataset for training the NN.

A practical inventory control problem is considered as an example, in which the risk objective and constraints can be evaluated or estimated with return variance.
An inventory control problem is dynamic in the long run.
For example, the wholesale prices are changing all the time, as well as the uncertainties of the market demands and supplier reliabilities.
How to adapt to these dynamics efficiently from a risk perspective is not only an academic concern, but also a business one. %concerned as well.
%Similar to an inventory control problem, many practical problems are dynamic. %the parameters involved in

\section{CONSTRAINED MARKOV DECISION PROCESSES AND RISK ESTIMATIONS}
In this section, 
firstly, we introduce the notations of constrained MDPs. 
Secondly, we present three types of law-invariant risk measures commonly studied in RL.
We show that, the three risk measures can be evaluated or estimated with the return variance, and any of the three risks can be the objective or a constraint in our study, as well as other functions of the mean and variance of return.
Thirdly, we review the return variance calculation method for a Markov process with a deterministic state-based reward.
Finally, we restate the SAT as an MDP homomorphism, which enables the variance calculation method for a Markov process with a stochastic transition-based reward.

\subsection{Constrained Markov Decision Processes} 
MDPs provide a framework for modeling sequential decision making process.
In this paper, we focus on infinite-horizon discrete-time MDPs. 
An MDP with a stochastic reward can be represented by
\[
\langle S, A, J, p, d, \mu , \gamma \rangle,
\]
in which
$S $ is a finite state space, and $X_t \in S$ represents the state at epoch $t \in \mathbb{N}$;
$A_x$ is the allowable action set for $x \in S$, $A = \bigcup_{x \in S}A_x$ is a finite action space, and $K_t \in A$ represents the action at epoch $t$;
$J$ is a bounded finite subset of $ \mathbb{R} $, and is the set of possible values of the immediate rewards, and denote $R_t$ the immediate reward at epoch $t$;
$ p $ is the transition probability, with $p(y \mid x, a) = \mathbb{P}(X_{t+1}=y \mid X_t = x, K_t = a)$ denotes the homogeneous transition probability;
$ d $ is the reward distribution, with $d(j \mid x, a, y) = \mathbb{P}(R_t=j \mid X_t = x, K_t = a, X_{t+1}=y)$ the probability that the immediate reward at time $ t $ is $ j $, given current state $ x $, action $ a $, and next state $ y $;
%Here we only consider $ d $ a discrete distribution with a finite support;
$\mu$ is the initial state distribution; and
$ \gamma \in (0,1)$ is the discount factor.
Given such an MDP, people concern the discounted total reward, or the return
\begin{equation}
\label{return}
\Phi = \sum_{t=1}^{+\infty} \gamma^{t-1} R_t.
\end{equation}
In most cases, the expected return is considered as the optimality criterion (objective).

A policy $\pi$ describes how to choose actions sequentially. 
An MDP with a (randomized) policy induces a Markov reward process.
% is tantamount to .
In an infinite horizon case, a stationary policy space is considered.
Randomized policy is often considered in constrained MDPs~\cite{altman1999constrained}. 
%Given an MDP with a randomized policy, the reward function is often naively simplified as well. 
In this study, we focus on the deterministic policy space $ \Pi $, which is easy-to-use in practice. 
An optimal policy with respect to a criterion refers to an induced Markov process which optimize the criterion.

%\color{blue}
In many practical SDM problems, a single objective might not suffice to describe the real cases. 
To deal with various goals, a natural way is to optimize one objective with constraints on other goals~\cite{altman1999constrained}.
In particular, since city logistics is characterized by multiple stakeholders with different objectives and constraints, we model such problems as constrained MDPs.
\color{black}
A constrained MDP is an MDP with a criterion composed of an objective and constraint(s).
%(\st{Given a policy $\pi \in \Pi$, an initial distribution $\mu$, and a discount factor $ \gamma $, we have the return $\Phi^{\pi}_{\mu} = \sum_{t=1}^{\infty} \gamma^{t-1} R_t$.
%To simplify the notation we denote it by $\Phi$.})
In this study, we consider risk-sensitive objective and constraints in the function space $ \{f \in \mathbb{R} \mid f= f(\mathbb{E}(\Phi), \mathbb{V}(\Phi))\} $, where the return $ \Phi $ refers to Equation~(\ref{return}).
%We focus on law-invariant risk measures~\cite{kusuoka2001law}, we generalize the transformation for settings mentioned above, in order to preserve the return distribution.
Next, we review the risk measures widely studied in RL.

\subsection{Risk Measures}
\label{secRisk}
In RL, uncertainty is studied from two perspectives. 
One is the \textit{external} uncertainty, which refers to the parameter uncertainty or disturbance.
When the model is unknown, its parameters are usually estimated first, and then the optimal solution is calculated with a model-based approach.
However, the parameter estimation depends on noisy data in practice, and the modeling errors may result in negative consequences.
In control theory, this problem is known as robust control.
Robust control methods consider the uncertain parameters within some compact sets, and optimize the expected return with the worst-case parameters, in order to achieve good robust performance and stability~\cite{nilim2005robust}.

The other refers to the \textit{inherent} (or \textit{internal}) uncertainty, which results from the stochastic nature of the process.
We claim that most, if not all, inherent risk measures depend on the reward sequence $ (R_t:t \in \{ 1, \cdots, N \}) $, %, i.e., a mapping from a sequence space of random variables to the real line.
and a inherent risk measure can be denoted by $ \rho: \mathbb{R}^N \rightarrow \mathbb{R} \cup \{ +\infty \} $, where $ N \in \mathbb{N}^{+} \cup \{+\infty\} $.
The inherent risk can be quantified by a dynamic measure or a law-invariant measure. 
Given a Markov process with a deterministic reward and a deterministic initial state, a sequence of risk measures $ (\rho_i: i \in \{1,\cdots, N\}) $, and denote the immediate reward at epoch $ t $ by $ R_t $, a dynamic measure can be denoted in general as
\[ 
R_1 + \rho_1(R_2 + \rho_2(R_3 +\rho_3(\cdots))), 
\]
%\[ 
%r(X_1, K_1) + \rho_1(r(X_2, K_2) + \rho_2(\cdots)), 
%\] ({\color{red} notation def.!})
which is sensitive to the order of the immediate rewards.
Dynamic measures are usually assumed to have a set of properties, such as Markov, monotonicity and coherence, which yields a time-consistent risk measure with a nested structure. 
For further information on the dynamic risk measure, see~\cite{ruszczynski2010risk}.

Given a discount factor $ \gamma $, a law-invariant~\cite{kusuoka2001law} measure in an infinite horizon is a functional $ \Psi $ on the return. %$ \Phi = \sum_{t=1}^{N} \gamma^{t-1} R_t $, where $ N \in \mathbb{N} \cup \{+\infty\} $.
%It is usually defined in terms of the total reward in a finite horizon  or the return in . %, and ignores the difference between immediate rewards received at different epochs. 
Three types of law-invariant risk have been widely studied in RL area. 

%\textit{Utility risk}: 
\subsubsection{Utility risk}
 %, in which the exponential utility function is concerned.
The original goal of a utility function is to represent the subjective preference~\cite{howard1972risk}. 
One classic example can be the ``St. Petersburg Paradox,'' which refers to a lottery with an infinite expected reward, but no one would put up an arbitrary high stake to play it, since the probability of obtaining an high enough reward is too small. %people only prefer to pay a small amount to play. 
%This problem is thoroughly studied in utility theory, and a recent study brought this idea to RL~\cite{Prashanth2015}.
Mathematically, a utility function $ \Psi: \mathbb{R} \rightarrow \mathbb{R} $ is a mapping from objective value space for all possible outcomes to subjective value space. 
A utility objective is usually in the form $ \Psi^{-1}\{\mathbb{E}[U(\Phi)]\} $, %consists of functionals of
where $ \Psi $ is a strictly increasing function.
Denoting the return variance by $ \mathbb{V}(\Phi) $, the most common used utility risk in RL is the exponential utility~\cite{chung1987discounted}
\[
\Psi(\Phi) = \beta^{-1}\log\{\mathbb{E}[\exp(\beta \Phi)]\},
\]
where $ \beta $ models a constant risk sensitivity that risk-averse when $ \beta < 0$.
This can be seen more clearly with the Taylor expansion of the utility
\[
\beta^{-1}\log\{\mathbb{E}[\exp(\beta \Phi)]\} = \mathbb{E}(\Phi) + \frac{\beta}{2}\mathbb{V}(\Phi) + \mathcal{O}(\beta^2).
\]

%\textit{Mean-variance risk}: 
\subsubsection{Mean-variance risk}
Another type of risk measure can be mean-variance risks.
The mean-variance risk measure is also known in finance as the modern portfolio theory~\cite{D.J.White1988a,doi:10.1287/opre.42.1.175,Mannor2011a}.
The mean-variance analysis aims at optimal return at a given level of risk, or the optimal risk at a given level of return.
In RL, several mean-variance models have been studied.
Denoting the return standard deviation by $ \sigma (\Phi) $, one model could be 
\[
\Psi(\Phi) = \mathbb{E}(\Phi) - k \sigma (\Phi),
\]
where $ k $ is a risk parameter, and when $ k > 0 $, it is a risk-averse objective.
This is the first mean-variance model for exploring inventory management related problems~\cite{lau1980newsboy}. 
The other model can be maximizing the expected return with a variance constraint, or minimizing the variance with an expected return constraint~\cite{choi2008mean}.
For a review on mean-variance risk, see~\cite{chiu2016supply}.

%\textit{Quantile-based risk}:
\subsubsection{Quantile-based risk}
The last type of risk measure used in practice refers to quantiles, which requires us to pay attention to discontinuities and intervals of quantile numbers.
A commonly used quantile-based risk measure is value at risk (VaR). 
VaR originates from finance. For a given portfolio, a loss threshold (target level) and a horizon, VaR concerns the probability that the loss on the portfolio exceeds the threshold over the time horizon. 
Mathematically, VaR can be defined as to find a policy which maximizes the smallest possible outcome with respect to a specified probability level. %maximizes the probability that the return is larger than or equal to a specified target (threshold)
In RL, the VaR objective can also be defined as to find a policy which maximizes the probability that the return is larger than or equal to a specified target (threshold)~\cite{Filar1995b,Wu1999a}. 
%See Section~\ref{SectionRisk} for the VaR definition.
Denote $F^{\pi}_{\Phi}$ as the return CDF from a policy $\pi$. 
In this paper, we consider two VaR problems~\cite{Filar1995b} in an infinite-horizon MDP. 
%Denote $\Pi$ as the stationary policy space. 
% For any target-percentile pair $(\tau, \alpha) \in \mathbb{R}\times [0,1]$
% For an MDP with an SAS-function, and an MDP with an SA-function derived from Equation (1), 

%\newtheorem{definition}{Problem}
\begin{problem}
\label{VaR_alpha}
	Given a quantile $\alpha \in [0,1]$, find the optimal threshold $\rho_{\alpha} =  \sup\{\tau \in \mathbb{R}\mid \mathbb{P}(\Phi > \tau) \ge \alpha, \pi \in \Pi\}=\sup\{\tau \in \mathbb{R}\mid F^{\pi}_{\Phi}(\tau) \le 1-\alpha, \pi \in \Pi\}$.
\end{problem}	

%This problem refers to the quantile function, i.e., $F^{\pi -1}_{\Phi}$.
%\footnotetext[1]{The equality is kept for the sake of the CDF semi-continuity in a finite time horizon MDP. To be mathematically precise, it should be $\sup_{\pi \in \Pi^N}\{\tau \in \mathbb{R} \mid F^{\pi}_{\Phi}(\tau) < 1-\alpha\}$.}

\begin{problem}
	Given a threshold $\tau \in \mathbb{R}$, find the optimal quantile $\eta_{\tau} = \sup\{\alpha \in [0,1]\mid F^{\pi}_{\Phi}(\tau) \le 1-\alpha, \pi \in \Pi\}$.
\end{problem}

%This problem concerns $F^{\pi}_{\Phi}$. 
Both VaR problems relate to  $P_{\Phi} = \inf\{F^{\pi}_{\Phi} \mid \pi \in \Pi\}$, %i.e., $P_{\Phi}(x)= \inf_{\pi \in \Pi} F^{\pi}_{\Phi}(x)$, for all $x \in \mathbb{R}$, so 
here we name it VaR function. %As will be illustrated below, when the horizon is short (Section 3), any point along $P_{\Phi}$ is $(\tau, 1-\eta_{\tau})$, and when the horizon is long (Section 4), and every (estimated) $F^{\pi}_{\Phi}$ is strictly increasing, the total reward distribution is estimated with a normal distribution, 
%any point along $P_{\Phi}$ is $(\rho_{\alpha}, 1-\eta_{\tau})$. %Since there exists a deterministic optimal policy for finite-horizon MDPs under VaR criteria \cite{Wu1999a}, we only consider the deterministic policy space. 
%The SA-function is commonly used in most MDP studies even considering risk (\cite{Filar1995b} for example) instead of the SAS-function. However, under the VaR criteria, if the original reward function is an SAS-function, the simplification will miss the optimality, i.e., neither the policy nor the VaR is optimal. In an MDP with an SAS-function, the simplified SA-function leads to the same optimal policy under the expected total reward criteria, but different optimal policies under the VaR criteria. Here we use a short-horizon inventory control problem to illustrate the effect of reward function on the optimal value under two criteria, and what is VaR about.
In a long or infinite horizon case, the return distribution can be estimated with a strictly increasing function~\cite{meyn2012markov}.
In this case, VaR can be considered as a study of the return distribution, since any point along $P_{\Phi}$ is (estimated) $(\rho_{\alpha}, 1-\eta_{\tau})$ with $\tau = \rho_{\alpha}$ or $\alpha = 1-\eta_{\tau}$. 
Therefore, both VaR problems refer to $P_{\Phi}$.
VaR is straightforward but hard to deal with since it is not a coherent risk measure~\cite{Riedel2004}.
In many cases, conditional VaR (also known as expected shortfall) is preferred over VaR since it is coherent~\cite{Artzner1998a}, i.e., it has some intuitively reasonable properties (convexity, for example). 
However, when the return can be assumed to be approximately normally distributed, VaR can be simply estimated with $ \mathbb{E}(\Phi)$ and $ \mathbb{V}(\Phi)$~\cite{Ma2017STT}.
%Even though VaR is hard to deal with, its intuitive meaning is straightforward.

In this paper, we focus on law-invariant risk measures on $ \Phi $.
The three law-invariant risk measures can be either calculated (mean-variance risk), estimated (utility risk), or estimated with assumption (quantile-based risk) with $ \mathbb{E} (\Phi) $ and $ \mathbb{V}(\Phi) $.
By virtue of the state-augmentation transformation (SAT), $ \mathbb{V}(\Phi) $ can be calculated in a Markov process with a stochastic reward, which allows risk evaluation in practical SDM problems. 
Next, we show how to calculate the return variance in a Markov process with a deterministic reward, and how the SAT enables the method to a Markov process with a stochastic reward.

\subsection{Variance Formula for Markov Processes}
\label{varCal}
As shown above, all law-invariant risk measures can be evaluated with the return variance.
To calculate the return variance, Sobel~\cite{sobel1982variance} presented the formula for the Markov process with a deterministic reward.

\begin{theorem} %~\cite{sobel1982variance} 
	Given an infinite-horizon Markov process $\langle S, r_{\pi}, p_{\pi}, \gamma \rangle$ with a finite state space $ S = \{1, \cdots, |S|\} $, a reward function $r_{\pi}$ deterministic state-based and bounded, and a discount factor $ \gamma \in (0,1)$. 
	Denote the transition matrix by $ P $, in which $ P(x,y) = p_{\pi}(y \mid x), x,y \in S $.
	Denote the conditional return expectation by $v_x = \mathbb{E}(\Phi  \mid  X_0 = x)$ for any deterministic initial state $x \in S$, and the conditional expectation vector by $v$. Similarly, denote the conditional return variance by $\psi_x = \mathbb{V}(\Phi \mid X_0 = x)$, and the conditional variance vector by $\psi$. 
	Let $\theta$ denote the vector whose $x$th component is $\theta_x = \sum_{y \in S}p_{\pi}(x,y) (r_{\pi}(x) + \gamma v_y)^2 - v^2_x$. 
	Then
	\begin{gather*}
		v = r_{\pi} + \gamma P v = (I - \gamma P)^{-1} r_{\pi}, \\ 
		\psi = \theta + \gamma^2 P \psi = (I - \gamma^2 P)^{-1} \theta.
	\end{gather*}
	\label{varUsed}
\end{theorem}
%Now with the aid of Theorem~\ref{theorem1982}, we can estimate the return distribution for the ergodic Markov reward process. 
Notice that the variance formula is for Markov processes with deterministic reward functions only.
How to apply the method to practical problems with stochastic rewards is the problem. %, and naively adopting the expected reward will lead to an incorrect risk value~\cite{Ma2017STT}. 
In next section, we review the SAT~\cite{shuai2019satAAAI} as an MDP homomorphism, which enables the variance formula in a Markov process with a stochastic reward.

\subsection{SAT Homomorphism}
\label{secHomo}
%For $ x, y \in S, a \in A_x $, a stochastic reward is often simplified to a deterministic one $ r(x,a) \in \mathbb{R}$, even in risk-sensitive studies~\cite{borkar2002q,shen2014risk,garcia2015comprehensive,gilbert2016quantile,huang2017risk,chow2017risk,berkenkamp2017safe}. 
%However, such a reward simplification changes the reward sequence $  (R_t)  $, and preserves $ \mathbb{E}(\Phi) $ only.
%In this case, we should apply a suitable SAT~\cite{shuai2019satAAAI} to transform the MDP (or Markov process) with $ d $ into one with $ r $, and preserve $  (R_t)  $ as well. 
In many practical risk-sensitive problems, the rewards of the Markov processes are stochastic, but many methods may require the reward in a determined form~\cite{borkar2002q,shen2014risk,garcia2015comprehensive,gilbert2016quantile,huang2017risk,chow2017risk,berkenkamp2017safe}.
To enable the method in those cases, we may use the SAT to transform the Markov process and preserve the reward sequence $ (R_t:t \in \{ 1, \cdots, N \}) $.
In this paper, the example is that the variance formula is for Markov processes with deterministic reward functions only.
%For Markov processes with (discrete) stochastic reward functions, we can transform it into ones with deterministic reward functions and preserve $ (R_t) $.
We restate the SAT as an MDP homomorphism.
Comparing with the original SAT theorem~\cite{shuai2019satAAAI}, the homomorphism version of SAT is on a more abstract level.
An MDP homomorphism is a formalism that captures an intuitive notion of specific equivalence between MDPs~\cite{ravindran2002model}.
In order to convert an MDP $ \mathcal{M} $ with $ d $ to $ \mathcal{M}^{\dagger} $ with $ r $ and preserve $ (R_t) $, we consider each ``situation'', which determines immediate reward, as an augmented state.
We can then attach each possible reward value to an augmented state in $ \mathcal{M}^{\dagger} $.
Formally, we define the SAT homomorphism as follows.

\begin{definition} [SAT, a homomorphism version]
	
	The SAT for MDPs is a homomorphism $ h $ from an MDP $ \mathcal{M} = \langle S, A, J, p, d, \mu \rangle $ to an MDP $ \mathcal{M}^{\dagger} = \langle S^{\dagger}, A, r, p^{\dagger}, \mu^{\dagger} \rangle $. 
	The state space $ S^{\dagger} = S^{\ddagger} \cup S_n $, where 
	$S^{\ddagger} = S^2 \times A \times J$, 
	$ S_n = \{s_{null,x}\}_{x \in S} $, %$ S_n = \{s_{null,1}, \cdots, s_{null,|S|}\} $, 
	and $ S_n \cap S^{\ddagger} = \emptyset $.
	%\[S^{\dagger} = S^2 \times A \times J . \]
	For $ x^{\dagger} = (x,a,y,i), y^{\dagger} = (y,a_y,z,j) \in S^{\ddagger} $, we have $ A_{x^{\dagger}} = A_y $, $ r(x^{\dagger}) = i $, and for $ a_y \in A_y $, $ p^{\dagger}(y^{\dagger} \mid (x^{\dagger}, a_y) = p^{\dagger}(y^{\dagger} \mid (s_{null,y}, a_y) = p(z \mid y,a_y) d(j \mid y,a_y,z) $; 
	for $ x^{\dagger} = s_{null,x} \in S_n $, we have $ A_{x^{\dagger}} = A_x $, $ r(x^{\dagger}) = 0 $, and $ \mu^{\dagger}(s_{null,x}) = \mu(x) $.
	
\end{definition}
We call $ \mathcal{M}^{\dagger} $ the homomorphic image of $ \mathcal{M} $ under $ h $.
For any policy $ \pi $ in $ \mathcal{M} $, there exists a policy $ \pi^{\dagger} $ in $ \mathcal{M}^{\dagger} $, such that the two processes share the same $ (R_t) $.
We define the mapping between the two policy spaces as a policy lift.

\begin{definition} [Policy lift]
	Let $ \mathcal{M}^{\dagger} $ be a homomorphic image of $ \mathcal{M} $ under $ h $.
	Let $ \pi $ be a stochastic policy in $ \mathcal{M} $.
	Then $ \pi $ lifted to $ \mathcal{M}^{\dagger} $ is the policy $ \pi^{\dagger} $ such that $ \pi^{\dagger}(a \mid x^{\dagger}) = \pi(a \mid y)$ for $ x^{\dagger} = (x,a,y,i) \in S^{\ddagger}$ and $ \pi^{\dagger}(a \mid x^{\dagger}) = \pi(a \mid x)$ for $ x^{\dagger} = s_{null,x} \in S_n$.
\end{definition}

Given an MDP with a policy, the randomness of the induced Markov reward process can be studied in its underlying probability space. %a sample path (trajectory)

\begin{definition} [Underlying probability space]
	Let $ (\Omega, \mathcal{F}, \mathcal{P}) $ be a probability space, and $ (E, \mathcal{B}) $ a measurable space with $ E = S \times J $.
	An induced Markov reward process can be represented by an $ (E, \mathcal{B}) $-valued stochastic process on $ (\Omega, \mathcal{F}, \mathcal{P}) $ with  a family $ (Y_t)_{t \in \mathbb{N}} $ of random variables $ Y_t: (\Omega, \mathcal{F}) \rightarrow (E, \mathcal{B}) $ for $ t \in \mathbb{N} $.
	$ (\Omega, \mathcal{F}, \mathcal{P}) $ is called the underlying probability space of the process $ (Y_t)_{t \in \mathbb{N}} $. %, while $ (E, \mathcal{B}) $ is a measurable state space.
	For all $ \omega \in \Omega $, the mapping $ Y(\cdot, \omega): t \in \mathbb{N} \rightarrow Y_t(\omega) \in E $ is called the trajectory of the process with respect to $ \omega $.
	The process $ (Y_t)_{t \in \mathbb{N}} $ is progressively measurable with respect to the filtration $ (\mathcal{F}_t)_{t \in \mathbb{N}} $. %, which is an increasing family of subalgebras of $ \mathcal{F} $.
\end{definition}

A homomorphism version of the SAT theorem is as follows, which claims that the probability measure on trajectories is preserved under $ h $.
Therefore, as a subsequence of sample path, the probability measure on $ (R_t)_{t \in \mathbb{N}} $ is preserved as well.
%A homomorphism version of the SAT theorem is as follows, which claims that, as a subsequence of trajectory, the probability measure on reward sequence $ (R_t)_{t \in \mathbb{N}} $ is preserved under $ h $.

\begin{theorem} [Probability measure preservation]
	\label{satUsed}
	Let $ \mathcal{M}^{\dagger} $ be an image of $ \mathcal{M} $ under homomorphism $ h $.
	Let $ \pi^{\dagger} $ be the stochastic policy lifted from $ \pi $.
	For the two processes $ \mathcal{M} $ with $ \pi $ and $ \mathcal{M}^{\dagger} $ with $ \pi^{\dagger} $, there exists a bijection $ f_{\Omega}: \Omega \rightarrow \Omega^{\dagger} $, such that
	for the underlying probability space $ (\Omega, \mathcal{F}, \mathcal{P}) $ for the first process, we have a sample path probability space $ (\Omega^{\dagger}, \{ f_{\Omega}(b): b \in \mathcal{F} \}, P^{\dagger}) $ for the second process, such that 
	for any $ t \in \mathbb{N} $, $ \mathcal{P}^{\dagger}(\{ f_{\Omega}(b): b \in \mathcal{F}_t \}) = \mathcal{P}(\{\mathcal{F}_t\})$.
	%$ P^{\dagger}(\{ f_{\Omega}(k): k \in b \}) = P(b) $.
\end{theorem}

%For $ x, y \in S, a \in A_x $,  
%However, such a reward simplification changes the reward sequence $  (R_t)  $, and preserves $ \mathbb{E}(\Phi) $ only.
%In this case, we should apply a suitable SAT~\cite{shuai2019satAAAI} to transform the MDP (or Markov process) with $ d $ into one with $ r $, and preserve $  (R_t)  $ as well. 
%\subsection{Risk Measures}
%In our problem, all functions on the mean and variance of return can be the objective and constraints.
%We consider VaR to be the risk-sensitive objective as an example.
%Having the variance formula and SAT, we can now estimate VaR.

%\begin{table}
%\caption{An Example of a Table}
%\label{table_example}
%\begin{center}
%\begin{tabular}{|c||c|}
%\hline
%One & Two\\
%\hline
%Three & Four\\
%\hline
%\end{tabular}
%\end{center}
%\end{table}

%\section{PRELIMINARIES AND NOTATIONS}
%In this section we present the notations for NN and MDPs with stochastic rewards, the variance formula for Markov processes with deterministic reward functions, the homomorphism version of the SAT for MDPs with stochasitc rewards, and two VaR problems as optional objectives or constraints. %, and common formulae for foreign exchange exposure (to be added). %with four types of reward functions and two policy space. %, which are concerned in the next section. introduce give 

\section{SEQUENTIAL DECISION MAKING SCHEME AND APPROXIMATOR}
In this section, we propose a scheme for solving risk-sensitive SDM problems with constraints in a dynamic environment.
And we consider neural network (NN) as an example of the approximator for the scheme.

\subsection{Sequential Decision Making Scheme}
A constrained, dynamic, risk-sensitive SDM problem can be considered as a process (environment) with a criterion (objective with constraints).
For scuh an SDM problem, we assume that its process is defined by a vector of $ k $ environmental features $ \mathbf{q_{e}} \in Q_e = F_1 \times \cdots \times F_k $, and denote it by $ \mathbf{q_{e,t}} $ at epoch $ t $.
%We assume that $ \mathbf{q_e} $ contains all related information for a decision maker in terms of the process.
We allow the components of $ \mathbf{q_e} $ to vary within some prespecified intervals.
Similarly, a targeted risk measure and risk constraints might be defined by a vector of parameters $ \mathbf{q_{p}} \in Q_p $, and and denote it by $ \mathbf{q_{p,t}} $ at epoch $ t $.
%If the risk objective and constraints need no parameter (variance for example), then $ Q_p = \emptyset $.
To allow the decision-making for dynamic risk concern, the components of $ \mathbf{q_p} $ are within some prespecified intervals as well.
For any given $ \mathbf{q_{e,t}} $ and $ \mathbf{q_{p,t}} $, denote the optimal risk by $ q_{o,t} \in Q_o $, and the optimal policy by $ \pi^*_t \in \Pi $. %A_p
In this case, an imagined decision maker can be mathematically denoted by $ \nu: Q_e \times Q_p \rightarrow Q_o \times \Pi $, i.e., a theoretically perfect decision maker consider all the environmental parameters, the objective, and possible constraints, and choose the optimal policy with respect to some targeted risk measure.
The goal is to train a NN to approximate $ \nu $. %for a specified risk objective and a set of risk constraints, with related parameters given from specified intervals.
%({\color{red} Here we set the outputs consisting of the optimal risk and policy, but later we focus on only one of them. Notations will be updated later.})

Usually, historical data is used in training an approximator.
However, at least two problems should be considered in practice. 
The first problem refers to information incompleteness. Most historical data contains no information on $ \mathbf{q_p} $ and (or) $ \pi^* $, and in many cases, even the information on $ \mathbf{q_e} $ is incomplete.
The second problem is optimality. In practice, the decision makers might prefer an easy-to-use policy than an optimal one, which is hard to determine since the practical problems could be different from the theoretical model diversely and subtly. %delicately
Therefore, we propose to evaluate or estimate risk measures with RL methods and train NN with a synthetic dataset.

\begin{figure*}[t]
%	\centering
	\includegraphics[trim={7cm 4cm 7cm 0},height = 1.1\textwidth, angle =-90]{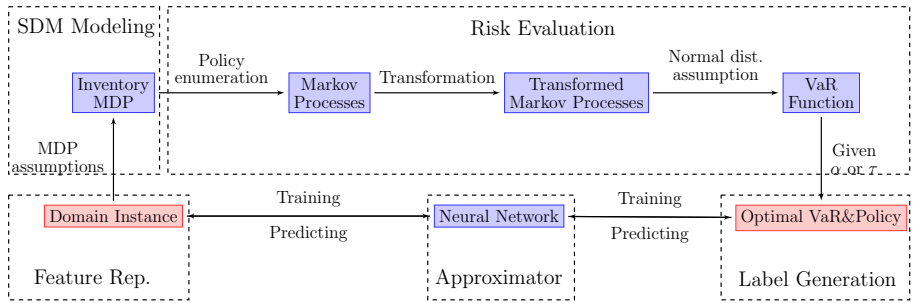}
	\caption{A dynamic risk evaluation scheme with NN and RL methods for optimal risk and policy in a sequential decision making problem.}
	\label{NNrisk1}
\end{figure*}

%\subsection{Scheme for Dynamic Risk-Sensitive SDM} %(how to do it in one way)
\label{SectionRisk}
We present a scheme in Figure~\ref{NNrisk1} to solve a constrained risk-sensitive SDM problem in a dynamic environment.
In the feature representation module, we define the SDM process with $ \mathbf{q_e} $. 
In the SDM modeling module, a mathematic model is chosen to represent the process and offer a platform for risk evaluation, and here we use MDP as the model. 
In the risk evaluation module, we calculate the optimal policy for a specified risk measure with risk constraints.
Here, for a given MDP with criterion (an objective with constraint(s), which can be risk-sensitive), firstly, we enumerate all deterministic policies to generate a Markov process.
Secondly, we implement SAT to transform the Markov process to one with a determined reward (See Section~\ref{secHomo}), in order to calculate the return variance.
Thirdly, we estimate the specified risk-measure(s) by the return variance (See Section~\ref{varCal}), and here we consider VaR as an example (See Section~\ref{secRisk}).
In the label generation module, the optimal VaR and policy is recorded and attached to the feature vector as a label. %we record optimal risk value and policy (decision) as the label of the process feature sequence.
Finally, a functional approximator, such as an NN or a linear combination of polynomial basis or Fourier basis, is trained with the labeled feature dataset. 
In this paper, we consider an inventory control problem as an example.
The feature set $ \mathbf{q_e} $ is fully described in Section~\ref{secExp}.
Briefly, the upper level of the scheme is for synthetic data generation (modeling and risk evaluation), %, which can be replaced by other methods.
and the lower level is for approximator training for dynamic SDM problems.
After each decision-making epoch, some parameters will be updated with the outcome (such as market demand distribution) and the involved external variables (such as wholesale price).
%Combining with the updated environmental parameters (such as the wholesale price), 
Suppose the updated parameters are still within the predefined intervals, the trained NN is able to output the estimated optimal policy and risk at the next decision-making epoch. %(or not?)
We present the algorithm for the synthetic dataset generation (upper level of the scheme) as follows.

\begin{algorithm}
	\small{
		\caption{Synthetic Dataset Generation Algorithm}\label{alg:Scheme}
		\hspace*{0.02in} {\bf Input:} 
		the size of training dataset $ |G| $, environmental feature space $ Q_e $, risk parameter space $ Q_p $. \\
		\hspace*{0.02in} {\bf Output:} 
		the training dataset $ G $.  
		\begin{algorithmic}[1]
%			\State Set the horizon $N^{\dagger} = N-1$;
%			\State Generate the state space $S^{\dagger} = S \times S$;
			\For{{\bfseries all} $ i \in \{1, \cdots, |G|\} $}
			\State Randomly generate a environmental feature vector $ \mathbf{q_{e}} \in  Q_e $ and a risk parameter vector $ \mathbf{q_{p}} \in Q_p $; 
%			\Comment{Consider VaR with constraints for example}
			\State Construct an MDP $ \mathcal{M} $ from $ \mathbf{q_{e}} $;
			    \For{{\bfseries all} $ \pi \in \Pi $}   \Comment{$ \Pi $ is the deterministic policy space}
%			        \State Get the Markov process $ \mathcal{M}^{\pi} $ from $ \mathcal{M} $ and $ \pi $;
			        \State Get the Markov process $ \mathcal{M}^{\pi} $ from $ \mathcal{M} $ and $ \pi $; \Comment{$ \mathcal{M}^{\pi} $ has a stochastic transition-based reward}
			        \State Get the transformed Markov process $ \mathcal{M}^{\pi \dagger} $ using the SAT; \Comment{$ \mathcal{M}^{\pi \dagger} $ has a deterministic state-based reward}
			        \State Calculate the return variance in $ \mathcal{M}^{\pi \dagger} $;
			        \State Estimate risk for $ \mathbf{q_{p}} $, and record the risk value and policy; %the VaR function
%			        \Comment{Multiple $ \mathbf{q_{p}} $ can be considered for efficiency}
			    \EndFor
			    \State Get the optimal risk value $ \rho^{*} $ and the corresponding optimal policy $ \pi^{*} $;
			    \State Record the $ i $-th sample point as $ (k_i, l_i) $, with $ k_i = (q_e, q_p) $ and $ l_i = (\rho^{*}, \pi^{*}) $);
			\EndFor
	\end{algorithmic}}
\end{algorithm}

\subsection{Approximator}
\label{ChNN}
A functional approximator needs training with a dataset to approximate the mapping from parameter space to risk and policy space.
In this study, We use neural network (NN), which is a universal function approximator~\cite{hornik1991approximation}. 
Other approximators, such as a linear combination of polynomial basis or Fourier basis, are not considered here.
%~\footnote{To be precise, a multilayer perceptron is a universal function approximator, as proven by the universal approximation theorem~\cite{hornik1991approximation}.}, 
%an NN needs to be trained and tested with a dataset.
%In this study, we generate a synthetic dataset to train and test an NN.
%See Section~\ref{dataGen} for details.
The definition of NN is as follows.

\begin{definition}[Feed forward neural network]
	Let $ Q \in \mathbb{N}$ be the number of layers, and $ N_0, N_1, \cdots, N_Q \in \mathbb{N}$ be the numbers of nodes of different layers. 
	Denote the activation function by $g: \mathbb{R} \rightarrow \mathbb{R} $. 
	For any $ q \in \{1, \cdots, Q\}, x \in \mathbb{R}$, denote the affine function by $ W_q(x) = A_q x + b_q $ for some $ A_q \in \mathbb{R}^{N_q \times N_{q-1}} $ and $ b_q \in \mathbb{R}^{N_q} $.
	For any $ i \in \{1, \cdots, N_q\}, j \in \{1, \cdots, N_{q-1}\} $, and $ q \in \{1, \cdots, Q\}$, the entry $ A_{q,i,j} $ of $ A_q $ denotes the weight of the edge from the $ i $-th node in the $ (q-1) $-th layer to the $ j $-th node in the $ q $-th layer.
	For any layer $ q \in \{1, \cdots, Q\} $, let $ W_q: \mathbb{R}^{N_{q-1}} \rightarrow \mathbb{R}^{N_q} $ be an affine function. 
	With $ H_q = g \circ W_q $ for $ q = 1, \cdots, Q-1 $, a function $ H:\mathbb{R}^{N_0} \rightarrow \mathbb{R}^{N_Q} $ defined as 
	\[
	H = W_Q \circ H_{Q-1} \circ \cdots \circ H_1 
	\]
	is called a feed forward neural network. 
%	The function $ g $ is called an activation function, which is applied componentwise. 
%	The integer $ Q $ denotes the number of layers. 
%	For any $ q \in \{0, \cdots, Q\} $, $ N_q $ denotes the number of nodes in the $ q $-th layer.	
	Since we use NN as an estimator of the decision maker $ \nu $, we denote the NN by $ \{\hat{\nu}^g_{\infty, N_0, N_Q}\} $ the set of neural networks mapping from $ \mathbb{R}^{N_0} $ to $ \mathbb{R}^{N_Q} $ with $ g $.
	
\end{definition}

In this study, for any given $ \mathbf{q_e} $, we construct an MDP model, and by virtue of the SAT~\cite{shuai2019satAAAI}, we can preserve the variance for a Markov process with a stochastic reward.
As claimed in Section~\ref{secRisk}, most inherent risk measures can be evaluated or estimated by return variance with or without some assumption.
We calculate $ q_o $ and $ \pi^* $ and attached them to $ \mathbf{q_e} $ and $ \mathbf{q_p} $ as labels.
Finally, we train an NN with the synthetic dataset.
In the next section, we present an inventory management problem as an example to show the implementation details.
%Next, we introduce constrained MDPs for modeling a practical SDM process.

\section{NUMERICAL EXPERIMENT}
\label{secExp}

Figure~\ref{NNrisk1} illustrates the scheme with NN and RL methods for a dynamic risk-sensitive SDM problem. %We take VaR with a given confidence level as an example. 
The scheme includes synthetic dataset generation (upper layer), approximator training and predicting (lower layer). 
To generate the dataset for training and testing, we define and randomly generate a domain set. %(contains many feature vector of inventory parameters) 
For each domain instance, we estimate the optimal VaR and the corresponding policy as follows.
Firstly, construct an inventory MDP under some predefined assumptions.
Secondly, enumerate all deterministic policies to achieve a set of Markov reward processes $\{\langle S, J_{\pi}, p_{\pi}, d_{\pi}, \mu\rangle\}_{\pi \in \Pi}$, and calculate the VaR function for the MDP with a stochastic reward. 
Thirdly, acquire the set of transformed Markov process $\{\langle S^{\dagger}, r^{\dagger}_{\pi^{\dagger}}, p^{\dagger}_{\pi^{\dagger}}, \mu^{\dagger}_{\pi^{\dagger}}\rangle\}_{\pi^{\dagger} \in \Pi^{\dagger}}$ by Theorem~\ref{satUsed}, with $ r^{\dagger}_{\pi^{\dagger}} $ being deterministic and state-based. 
Fourthly, estimate the mean and variance vectors $ v, \psi $ by Theorem~\ref{varUsed}. %~\footnote{Randomized policy are also allowed, and should be considered in constrained problems.}
Assuming the return is approximately normally distributed, we have the estimated return distributions for all $\langle S^{\dagger}, r^{\dagger}_{\pi^{\dagger}}, p^{\dagger}_{\pi^{\dagger}}, \mu^{\dagger}_{\pi^{\dagger}}\rangle$ to calculate the VaR function for the inventory MDP.
Finally, for the risk sensitivity parameters in the domain instance, calculate risk-sensitive objective and constraints, and record $ \pi^* $. %for the specified confidence level
%Next step is to acquire the VaR value for the given risk sensitivity parameters, here we calculate the optimal quantiles $ \tau $ for all given confidence levels $ \alpha $, and record the corresponding policies.
%Now we have the data set to train and test the NN, %For different hyperparameter sets, we follow the same procedure to generate the data set
%the inputs (related inventory parameters and confidence levels) and the outputs (optimal quantiles and corresponding policies) of the function we want, we can train the NN with this data set. 
%Notice that we can generate the data set as big as we want.
With the training dataset, we tune, train and validate an NN properly, and use it to predict $ q_o $ and $ \pi^* $ for any $ \mathbf{q_e} $ and $ \mathbf{q_p} $. %inventory control problems with the same hyperparameters but different domain instances.

\subsection{An Inventory Management Example}
%\color{blue}
In the management of sustainable city logistics, both costs and risks are required to be
simultaneously evaluated, and these are often conflicting~\cite{taniguchi2012emerging}. 
Moreover, considering the changing environment, the solution should be sensitive to the dynamics of the model parameters $ Q_e \times Q_p $.
As far as we know, most decision-making methods lack a holistic integrated vision, and do not attempt to safeguard sustainable city logistics security at different levels, thus causing unnecessarily high costs and disturbances. 
To shift from the risk transfer and toleration towards a risk-sensitive control, the proposed scheme in Figure~\ref{NNrisk1} can be regarded as a solution in this aspect.
\color{black}
As a proof of principle for the applicability of the proposed scheme, we apply it with NN to a practical inventory control process as an example, in which three factors are modeled: multiple sourcing, supplier reliability, and backlog.
%\begin{itemize}
%\item 
%%\item transportation
%\item 
%\item 
%%\item fast delivery (pending)
%\end{itemize}
%Before amplifying further, i
It is worth noting that MDP is a discrete-time process, so the output optimal policy is a periodic-review strategy, which has its limitations comparing with its aperiodic-review counterparts.
Furthermore, some assumptions are made to keep the model small.
Hence, this inventory control SDM model is theoretical, and in many cases a more accurate estimation can be achieved through simulation with respect to complicated cases.

%\textit{Multiple sourcing}: 
\subsubsection{Multiple sourcing}
Multiple sourcing (multisourcing) is a strategy that blends services from the optimal set of internal and external suppliers in the pursuit of business goals~\cite{cohen2006multisourcing}.
We consider two suppliers $U_1$ and $U_2$, and a retailer $R$. % for example.
The scenario involves a multinational corporation ($R$) which functions in the supply chain as a retailer. % (such as Costco in Canada).
For a targeted product, this corporation has two suppliers $U_1$ and $U_2$ (in US and in India, for example).
The supplier $U_1$ in US is close and reliable, the other supplier $U_2$ is far and unreliable but with a lower wholesale price.
%Here we only consider the wholesale prices, reliability and transportation as the factors influenced by suppliers.

%\paragraph{Transportation} 

%\textit{Supplier reliability}: 
\subsubsection{Supplier reliability}
Supplier (sourcing) reliability can be expressed in terms of quantity, quality or timing of orders due to multiple factors, such as equipment breakdowns, material shortages, warehouse capacity constraints, price inflations, strikes, embargoes, and political crises~\cite{Mohebbi2004}.
In the framework of MDP, one way to model these uncertainties is to take the status of a supplier as a Boolean variable, which represents whether the supplier is ``available'' or ``unavailable'', and it can be modeled as a two-state Markov process~\cite{ahiska2013markov}. %~\footnote{In the last version, we use one state component to represent this.}.
In order to simplify the state representation, we assume that the unreliable supplier $ U_2 $ is available with a probability $ \beta_1 $, and the retailer will be compensated with an amount $ p_s $ if the supplier with an order is unreliable at the current epoch.
The probability $ \beta_1 $ can be estimated using historical data.

Transportation can be another source of unreliability. It emphasizes the long delivery time which mainly results from a long physical distance.
This uncertainty can be also a critical factor at times.
In our model, we assume $U_1$ is reliable and its transportation lead time can be ignored (for example, a day when one epoch is a month). 
%We assume that this producing and delivery time (lead time?) .
Assume $U_2$ offers a lower price but unreliable.
Since it is far from $ R $, its lead time is one period.
To simplify the model, we consider the transportation factor within the supplier reliability.
%We assumpe the transportation (delivery) fails with a probability $\beta_1$, after which the retailer will be compensated and the order will be canceled. %, and the order will be returned to the supplier.

%\textit{Backlog and lost sale}: 
\subsubsection{Backlog and lost sale}
For each unsatisfied unit of demand, a backlog or a lost sale occurs.
Suppose the probabilities of backlog and lost sale are $ \beta_2 $ and $ 1-\beta_2 $, respectively.
The probability $ \beta_2 $ can be estimated using historical data.
We assume that when a backlog happens, the product will be sold at a lower price $ p_r - c_b $ (which can be due to from fast delivery or other reasons), where $ p_r $ is the retail price.
A lost sale induces a cost $ c_l $.
Rewards for both cases will be calculated at the current epoch, in order to keep the state representation small.
The backlog and lost sale instances are associated with market demand.
To keep the stochastic reward function simple, finite possible demands are considered (i.e., a finite support for the reward function).
For example, when the warehouse capacity is four units and the maximum demand is six units, then the possible maximum amount of backlogs or lost sales is two units.
In the MDP model, the reward and cost from these two units will be received at a discount at the current epoch.

%\paragraph{Fast delivery}
%Fast delivery (from the farther supplier with zero delivery time and no failure) can be offered with a cost $ Cd $.
%But this concern will augment the action space (as one component in the action representation), so we ignore it currently.

\begin{figure}[t]
	\centering
	\includegraphics[width = \textwidth]{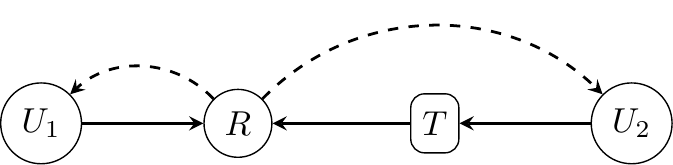}
	\caption{The product (solid) and order (dashed) flows between the retailer $ R $ and the two suppliers $ U_1 $ and $ U_2 $. The letter $T$ denotes transportation. The transportation lead time of $ U_1 $ is negligible, hence ignored.}
	\label{inventoryFlow_new}
\end{figure}

To simplify the model, other factors, such as fast delivery, order replacement and foreign currency exchange, have not been considered.
The inventory control process is illustrated in Figure~\ref{inventoryFlow_new}, in which the solid arrows represent the product flow, and the dashed arrows represent the order flow. %~\footnote{The two words``order flow'' and ``product flow'' are fabricated.}
Next, we construct an MDP for this inventory control problem.

\subsection{Inventory MDP}
Here we use an infinite-horizon MDP with finite state and action space to model the inventory control process. The related parameters are given in Table~\ref{inventoryParamList_new}.
For a set of specified parameters, we construct the MDP as follows.
For time $t \in \mathbb{N}$, denote the inventory level by $I_t$, and the number of items in transit $J_t$. 
Given the warehouse capacity $ M $, the state is $S_t=(I_t, J_t)$, where $I_t, J_t \in \{0, \cdots, M\}$. 
The action is $K_t=(k_{t,1}, k_{t,2})$, where $k_{t,1}, k_{t,2} \in \{0, \cdots, M\}$ are the order quantities with the two suppliers. 
Assume that the maximum demand at retailer could be double of the warehouse capacity, i.e., $ \supp(f_D) = \{0, \cdots, 2M\} $, and the demand is $ d_t \in \supp(f_D) $ with probability $ f_D(d_t) $. 
Given $\beta_1$ and $\beta_2$ to represnet supplier availability and backlog probabilities, respectively, the reward and transition probability can be formulated. % as
%\[
%R_t \sim D\{r[(i_t,j_t), (k_{t,1}, k_{t,2}), (max(i_t+j_t+k_{t,1}-d_t,0),k_{t,2})]\} 
%\] 
For example, at time $t$, given the state $S_t=(i_t,j_t)$, the action $K_t=(k_{t,1}, k_{t,2})$, the demand $d_t \le i_t+j_t+k_{t,1}$, $ U_2 $ being available currently, and the next state $S_{t+1}=(i_{t+1}, k_{t,2})$, the reward is %can be calculated as%, no fast delivery

\begin{flushleft} %= D\{r[(i_t,j_t), (k_{t,1}, k_{t,2}), (max(i_t+j_t+k_{t,1}-d_t,0),k_{t,2})]\} $
	$R_t =p_r \cdot d_t - c_f \cdot (\mathds{1}_{[k_{t,1} > 0]} +  \mathds{1}_{[k_{t,2} > 0]}) - (c_1 \cdot k_{t,1} + c_2 \cdot k_{t,2}) - c_h \cdot i_t,$ %+u_{t,1}+u_{t,12}
\end{flushleft}
where $ d_t = i_t+j_t+k_{t,1} -i_{t+1}$. 
Then, the transition probability is given by
\begin{flushleft}
	$\mathbb{P}(S_{t+1}, \text{ `$ U_2 $  is available'} \mid S_t, K_t)=f_D(d_t) \cdot \beta_1,$
\end{flushleft}

Consider another example, where $ U_2 $ is unavailable.
For the backlogging of one unit and the lost sale of one unit, the reward is %can be calculated as%, no fast delivery
\begin{flushleft} %= D\{r[(i_t,j_t), (k_{t,1}, k_{t,2}), (max(i_t+j_t+k_{t,1}-d_t,0),k_{t,2})]\} $
	$R_t =p_r \cdot (d_t-1) - c_f \cdot (\mathds{1}_{[k_{t,1} > 0]} +  \mathds{1}_{[k_{t,2} > 0]}) - (c_1 \cdot k_{t,1} + c_2 \cdot k_{t,2}) - c_h \cdot i_t + p_s \cdot k_{t,2} -c_b-c_l,$ %+u_{t,1}+u_{t,12}
\end{flushleft}
and the transition probability is
\begin{flushleft}
	$\mathbb{P}(S_{t+1}, \text{ `$ U_2 $  is unavailable', `one backlog', `one lost sale'} \mid S_t, K_t)=f_D(d_t) \cdot (1-\beta_1) \cdot \beta_2 \cdot (1-\beta_2).$ %`transformation fails'
\end{flushleft}

Next, we consider a VaR objective with a risk-sensitive constraint, and generate the synthetic training data for NN.

\begin{table}[t]
	\caption{Inventory control parameter list}
	\label{inventoryParamList_new}
	\begin{center}
		\begin{tabular}{|l||l|}
			\hline
			%$N$ & Time horizon \\ \hline
			$M$ & warehouse capacity \\  %(or $(m_i, M_i)$ instead) 
			$\gamma$ & Discount factor \\ 
			$\mu$ & Initial state distribution \\ \hline
			$p_r$ & Retail price \\
			$p_s$ & Penalty on supplier's unavailability \\
			$c_1, c_2$ & Wholesale price from $ U_1 $ and $ U_2 $ \\
			$c_b$ & Backlog cost \\
			%$Pt$ & Transportation failure compensation \\
			$ c_l $ & Lost sale cost, which could be a dependent variable \\
			%$ Cd $ & Fast delivery cost \\
			$c_f$ & Fixed ordering cost \\
			$c_h$  & Holding cost \\ \hline
			$f_D$ & Demand distribution function (PMF)\\
			%$\beta_1$ & Transportation failure probability\\
			$\beta_1$ & Supplier availability probability\\
			$\beta_2$ & Backlog probability\\
			$\alpha$ & Risk sensitivity parameter (for VaR)\\
			%  $p_d$ & Faster delivery probability \\
			%$w_0, w_1$ & Source status transition probabilities \\
			\hline
		\end{tabular}
	\end{center}
\end{table}

\subsection{Dataset Generation}
\label{dataGen}
In this paper, we consider the VaR Problem~\ref{VaR_alpha} as the objective with a constraint $ \mathbb{E}(\Phi) / \mathbb{V}(\Phi) > q $ as an example, in which $ q $ is a prespecified constant. %(see Section~\ref{secExp})
The intuitive meaning of this constraint is that the earning per unit of risk (variance) should be larger than a threshold $ q $.
We artificially generate a dataset for the inventory control process with an objective of minimizing VaR with a confidence level $ \alpha$ and a constraint $ \mathbb{E}(\Phi) / \mathbb{V}(\Phi) > q $, in which $ q $ is a constant.

%We first train an NN for a simplified invnetory control problem with a VaR objective.
%The risk objective can be a (exponential) utility [ref], a function of mean and variance of the return [ref], or VaR. 
%We consider VaR as an example. 
%VaR takes either a confidence level or a threshold as the risk sensitivity parameter. Here we consider VaR $ \rho_{\alpha} $ with a given $ \alpha $ as the objective.
There are multiple ways to define a mapping from parameter space to risk and policy space. Here we define it as follows. 
Define the domain set $ \mathcal{K} $, with each instance $ k \in \mathcal{K} $ refering to a vector of inventory control parameters (retail price, fixed ordering cost, etc.). 
Define the label set $ \mathcal{L} $, with each instance $ l \in \mathcal{L}$ refering to a vector of labels (optimal VaR, optimal policy, etc.).
A dataset $ G = \{(k_1, l_1), \cdots, (k_m, l_m)\} $ is a finite set of pairs in $ \mathcal{K} \times \mathcal{L} $, i.e., a set of labeled domain instances. 

It worth noting that, to construct an NN one needs to predefine certain parameters, including but not limit to $ Q $, $ \{N_i\} $ and $ g $ (See Section~\ref{ChNN} for details). 
The predefined parameters, which determine the network structure and how the NN is trained, are called hyperparameters.
%Hyperparameters are set before training (optimizing $ \{A_q\} $ and $ \{b_q\} $).
A fine-tuned NN with a large enough dataset should be able to make a satisfactory prediction.
%Our method aims at providing an NN ~~~optimal VaR and policy for a specified inventory control problem, which is represented by a parameter set and a risk sensitivity parameter.
%The inputs of the function we want to use NN to esitmate are . 
For our inventory control problem, all related parameters are set and sampled as follows. 

\textit{Hyperparameters}: 
A domain instance includes all variable values except for warehouse capacity $ M$, discount factor $\gamma$, and initial inventory distribution $\mu$, which are considered as hyperparameters.
We set $ M = 3 $, and the inventory level space $ S = \{0,\cdots, M\} $, and $ \gamma = 0.95 $. %In the future we can take $ \gamma $ as a random variable. %and it is better to keep it close to 1 (greater than or equal to 0.90, intuitively).
We set $ \mu((0,0)) = 1 $, i.e., at the beginning the inventory level and the item amount in transit are both zero. %but it can be dynamic.
The latter two parameters are fixed only to simplify the problem.

\textit{Revenue related parameters}: %Price related
Set the retail price $ p_r \in P_r = [6,10]$, and for generating the dataset, we take it as a random variable, whose distribution is $ U(6,10)$~\footnote{A uniform distribution with a support $ [6,10] $.}.
Similarly, set the penalty on supplier's unavailability $ p_s \in P_s = [0,2] $ with $ p_s \sim U(0,2) $,
the wholesale prices from $ U_1 $ and $ U_2 $ are $ c_1 \in C_1 = [4,6]$ with $ c_1 \sim U(4,6) $ and $ c_2 \in C_2 = [1,4] $ with $ c_2 \sim U(1,4) $,
the backlog cost $ c_b \in C_b = [0,2] $ with $ c_b \sim U(0,2) $,
the fixed ordering cost (for both supplier) $ c_f \in C_f = [0,2]$ with $ c_f \sim U(0,2) $, and
the holding cost $ c_h \in C_h = [0,2] $ with $ c_h \sim U(0,2) $.
For brevity, we set the lost sale cost $ c_l = p_r - c_2$.

For probabilistic parameters, we set the supplier availability probability $ \beta_1 \in [0,1]$ with $ \beta_1 \sim U(0.8,1) $,
the backlog probability $ \beta_2 \in [0,1]$ with $ \beta_2 \sim U(0,1) $, and
the risk sensitivity parameter $ \alpha \in [0,1]$ with $ \alpha \sim U(0,1) $.
For the demand distribution vector $ v_D \in [0,1]^{2M+1}$, we sample it uniformly from a $ 2M $-simplex. %~\footnote{When the backlog factor is ignored, the demand distribution should be sampled from a $ M $-simplex. For simplex, see~\url{https://en.wikipedia.org/wiki/Simplex}.}.
After defining the domain features, labels are calculated, which include the optimal risk and policy.

We denote a policy list (vector) by $ L_p $, in which each item represents an allowable policy.
An NN will be trained as an approximator of the function, whose inputs represent the inventory parameters and $ \alpha $, and outputs represent $ \rho_{\alpha} $ and the corresponding policy (denoted by a Boolean vector or an index), i.e., the domain set $ \mathcal{K} = P_r \times P_s \times C_1 \times C_2 \times C_b \times C_f \times C_h \times [0.8,1] \times [0,1]^{M+3} $, and the label set $ \mathcal{L} = \mathbb{R} \times \{0,1\}^{|L_p|} $ in general. 
Next, we train an NN with the synthetic dataset and show the validity of the proposed scheme.
%In our example, one possible labeled instance could be $ j = (k,l)$, where $k=(6, 4, 1, 2, 0.25, 0.5, 0.25, 0.9), l=( 65, [0,0,0,1,0,0]) $. The domain instance $ k $ represents that, 
%the retail price $ c_r =6 $, 
%the fixed ordering cost $ c_f =4 $, 
%the holding cost $ c_h =1 $, 
%the wholesale price $ c_p =2 $, 
%the demand distribution $ \mathbb{D}_d =[0.25, 0.5,0.25] $, and
%the risk sensitivity parameter (confidence level) $ \alpha =0.9 $.
%The label instance $ l $ represents that, given the domain instance $ k $,
%the optimal VaR $ \rho_{\alpha} =65 $, and
%the optimal policy index is 4. 

%Since the outputs (risk value and policy) are in different variable types (numeric and categorical), we deal with them in two separate NNs.
%For risk evaluation, set $ N_0=14, N_Q =1 $, a network $ NN^g_{w, 14, 1}: \mathcal{K} \rightarrow \mathcal{L}$ can be trained (by adjusting $ \{A_q\} $ and $ \{b_q\} $, and $ g,w $ are determined by trial and error) and validated with a dataset. The number of non-zero weights $ w $ and $ g $ are to be determined. Given all hyperparameters determined, the network $ NN^g_{w, 14, 1} $ is trained with a training dataset $ G_t \subset G $, i.e., $ NN^g_{w, 14, 1} = NN^g_{w, 14, 1}(J_t) $~\footnote{To choose the hyperparameters is an empirical process, and the training procedure will be explained when the NN is fine-tuned.}. 
%Similarly, for optimal policy, set $ N_0=14, N_Q = |L_p| $.

\begin{figure}
	\centering
	\includegraphics[width=.95\linewidth]{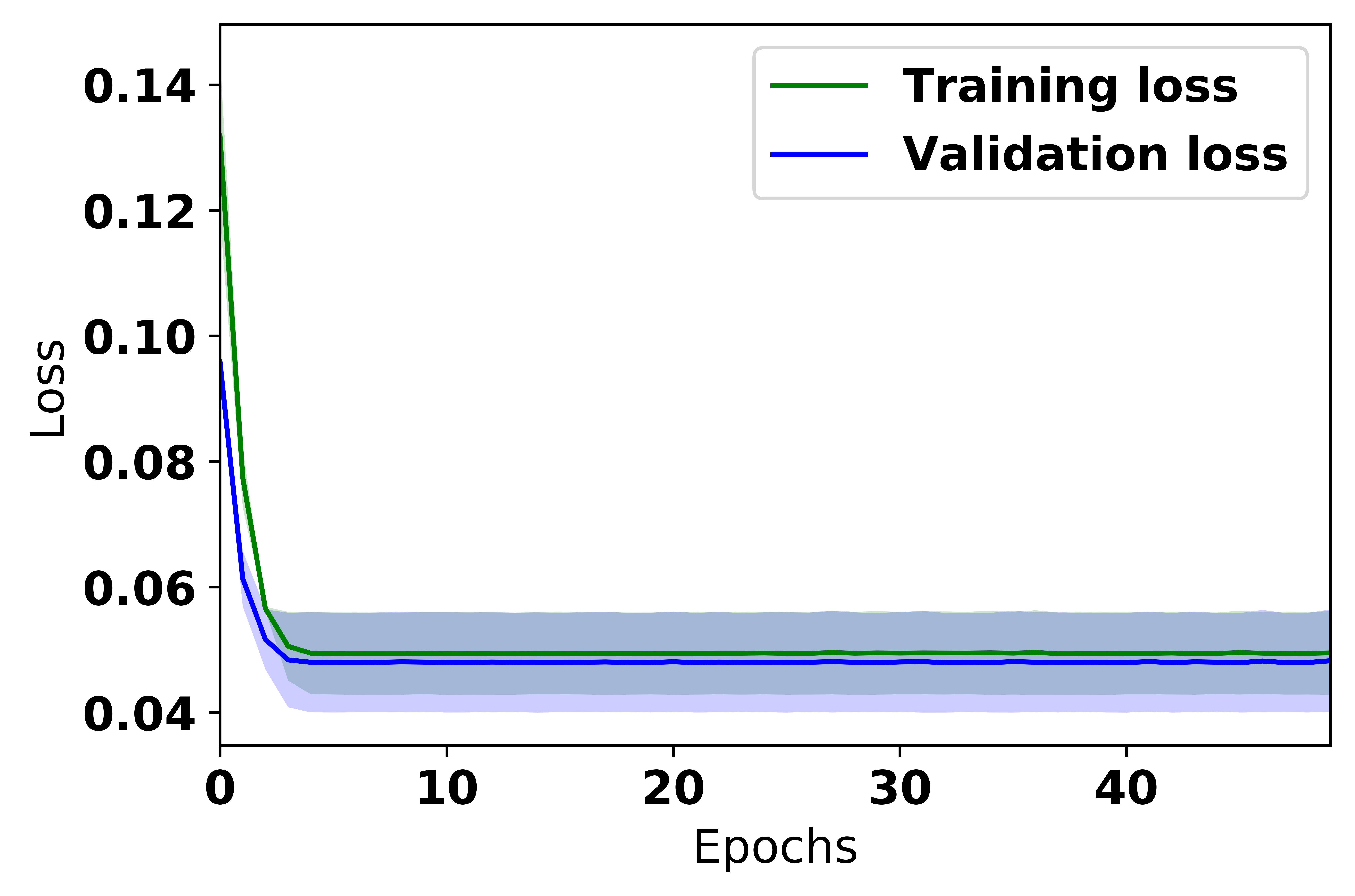}
	
	\caption{The loss for training/validating a 3-layer network in 50 epochs.}
	\label{loss}
\end{figure}
\subsection{Numerical Result}
Since VaR is numeric and (deterministic) inventory policy is categorical, we consider both of them to be numeric. %, and the policy space suffers a combinatorial explosion
%Furthermore, real data usually does not include VaR, which can only be estimated~\footnote{For an inventory control problem, I suppose the accessible data does not have a number to describe how risky the chosen policy is.}.
Considering the setting given in Section~\ref{dataGen}, we set a network $ NN^g_{w, 17, 7}: \mathcal{K} \rightarrow \mathbb{R}^7$.
%There are multiple ways to define the output of the NN
%For a state $ s \in S $, our problem can be considered as a multi-class classification problem in an NN, i.e., which action $ a \in A_s $ is optimal.
%For the toy example above, there are six policies in total. 
%Therefore, it is suitable to consider the output as a vector $ [p_i]_{i \in \{1, \cdots, 6\}} $, in which $ p_i $ represents the probability that the $ i $-th policy is the optimal.
Now we use Keras to construct, train and validate an NN for this small inventory problem. 
Firstly, we try a 3-layer NN. % for this small inventory problem.
We set $ Q=3$, the numbers of nodes $N_0 =17, N_1 =12, N_2=8, N_3 = 7 $.
We set the \textit{relu} function and the linear function %~\footnote{For relu and softmax functions, see \url{https://en.wikipedia.org/wiki/Rectifier_(neural_networks)} and \url{https://en.wikipedia.org/wiki/Softmax_function}.} 
as the activation functions for the hidden and output layers, respectively. 
Set the \textit{mean squared error} as the loss function, and \textit{adam} as the optimizer.
Given that all hyperparameters are determined, the network is trained with a training dataset $ G_t \subset G $.
We train the NN 50 times for the means and variances of the loss values at different epochs.
%In each training phase, the training dataset is went through 50 times.
We set the batch size to be 50 for the batch gradient descent. 

%To determine the size of the training set, we try 10 different sizes from $ 10^5 $ to $ 10^6 $, with a fixed size of the validation data set $ 10^3 $, i.e., try the data sets $ J_M = \{(k_1, l_1), \cdots, (k_{M+10^3}, l_{M+10^3})\}, M \in \{10^5, 2\cdot10^5, \cdots, 10^6\} $.

%To determine the size of the training set, we try 10 different sizes from $ 10^5 $ to $ 10^6 $, with a fixed size of the validation data set $ 10^3 $, i.e., try the data sets $ J_M = \{(k_1, l_1), \cdots, (k_{M+10^3}, l_{M+10^3})\}, M \in \{10^5, 2\cdot10^5, \cdots, 10^6\} $.
%\begin{figure*}
%	\centering
%	\begin{subfigure}{.5\textwidth}
%		\centering
%		\includegraphics[width=.9\linewidth]{keras_result_acc_dataSize_new}
%		%  \caption{The loss curves for training/testing\\in 10 epoches.}
%		%  \label{fig:sub1}
%	\end{subfigure}%
%	\begin{subfigure}{.5\textwidth}
%		\centering
%		\includegraphics[width=.9\linewidth]{keras_result_loss_dataSize_new}
%		%  \caption{.}
%		%  \label{fig:sub2}
%	\end{subfigure}
%	\caption{The loss/accuracy curves for a 2-layer network with the sizes from $ 10^5 $ to $ 10^6 $ of the training data sets, and a fixed size $ 10^3 $ of the validation data sets.} 
%	\label{Keras_datasize}
%\end{figure*}
%Figure~\ref{Keras_datasize} illustrates that, in general, the performance of the NN is better with a larger data set.
Since it is a regression problem, we measure the loss by the regression metrics, such as mean absolute error or mean squared error (MSE).
In Figure~\ref{loss}, it shows that the result with the training data size $ 10^5 $ %is already good---from Figure~\ref{loss}, we can see that it 
converges within 10 epochs, with a hit rate around $ 95\% $ in both training and validation phases. The error regions represent the standard deviations of means along the epoch axis. 

\section{RELATED WORK}
%SDM;SDM with risk; SDM in dynamic environment1; SDM with constraint2
% % % % % % % % % % % %
The SDM problems considering risk, dynamic environment, and constraint are usually studied separately.
% 0this paper proposed valuation functions and study exponential utility function with model-free RL algorithms.
Besides the works reviewed in Section~\ref{secRisk}, Shen~\cite{shenyun2015} generalized risk measures to the valuation functions. 
The author applied a set of valuation functions, derived some model-free risk-sensitive reinforcement learning algorithms, and presented a risk control example in simulated algorithmic trading of stocks.
For SDMs in dynamic environments, 
Hadoux~\cite{hadoux2015markovian} proposed a new model named Hidden Semi-Markov-Mode Markov Decision Process (HS3MDP), which represented non-stationary problems whose dynamics evolved among a finite set of contexts. 
The author adapted the Partially Observable Monte-Carlo Planning (POMCP) algorithm to HS3MDPs in order to efficiently solve those problems. 
The POMCP algorithm used a black-box environment simulator and a particle filter to approximate a belief
state. 
The simulator relaxed the model-based requirement, and each filter particle represented a state of the POMDP being solved.
%Therefore, with an infinite-sized filter, the particle repartition would exactly match
%the belief state of the POMDP.
%Moreover,
%it does not require to reflect exactly the real environment, at the cost, of course, of
%a less optimal solution.
For different types of dynamic environment, the author compared a regret-based method with its Markov counterpart~\cite{choi2001solving}.
In the regret-based method, the agent was involved in a two-players repeated game, where two agents (the player and the opponent, which can be the environment) chose an action to play, got a feedback, and repeated the game.
% for dynamic environment
%Those methods are as diverse as the different types of non-stationarity. Among
%them, two types of methods are prominent: regret-based and Markov methods.
%There exists several methods minimizing the regret under different assumptions
%(see, for instance, (Nisan et al., 2007; Cesa-Bianchi and Lugosi, 2006; Bubeck and
%Cesa-Bianchi, 2012)).
%
%Choi et al. proposed the Hidden-Mode Markov Decision Process (HM-MDP)
%model to formalize this subclass of non-stationary problems (2001). The environ-
%mental changes are limited to a fixed and known number n of modes. Each mode
%represents a possible stationary environment, formalized as an MDP.
%

For SDMs with constraint(s),
% 2a good paper can be referred in terms of LR and other parts. CMDP, constraints, and CVaR.
% chapter structure: Introduction (...; Chapter Contribution; Chapter Organization)
Chow~\cite{chow2017risk} investigated the variability of stochastic rewards and the robustness to modeling errors. 
The author analyzed a unifying planning framework with coherent risk measures (such as CVaR), which was robust to inherent uncertainties and modeling errors, and output a time-consistent policy.
A scalable approximate value-iteration algorithm on an augmented state space was developed for large scale problems with a data-driven setup.
%In addition, we discover an
%interesting relationship between the CVaR risk of total cost and the worst-case expected cost under adversarial
%model perturbations, which leads to a robustness framework that is significantly less conservative than robust-
%MDP.
The author proposed novel policy gradient and actor-critic algorithms for CVaR-constrained and chance-constrained optimization in MDPs. 
Furthermore, the author proposed a framework for risk-averse model predictive control, where the risk was time-consistent and Markovian. 
%In Chapter 5, we study a dynamic programming approach to stochastic optimal control problems with
%dynamic, time-consistent (in particular Markov) risk constraints. 
% % % % % % % % % % % % % %
% 2this paper is for constraint in RL based on CMDP framework with constraints on expected cumulative costs. Our approach hinges on a novel Lyapunov method.
%% some works dealt safety with constraints on expected cumulative cost.
In the framework of CMDP~\cite{altman1999constrained}, Chow et al.~\cite{chow2018lyapunov} derived two LP-based algorithms---safe policy iteration and safe value iteration---for problems with constraints on expected cumulative costs. 
The algorithms hinged on a novel Lyapunov method, which constructed Lyapunov functions to provide an effective way to guarantee the global safety of a policy during training via a set of local, linear constraints. 
For unknown environment models and large state/action spaces, the authors proposed two scalable safe RL algorithms: safe DQN, an off-policy fitted Q−iteration method, and safe DPI, an approximate policy iteration method. 
% % % % % % % % % % % % % % % % % % % % % % % % % % %
% 02this paper study risk as a constraint with the expectation of the times of that the state is in a set of error states no greater than a predefined threshold.
Another way to consider risk in an SDM problem is to avoid some dangerous system states.
Geibel and Wysotzki~\cite{geibel2005risk} defined the risk with respect to a policy as the probability of entering such states, and set the constraint as the probability being smaller than a predefined threshold.
The authors presented a model-free RL algorithm with a deterministic policy space.
The algorithm was based on weighting the original value function and the risk. 
The weight parameter was adapted to find a feasible solution for the constrained problem, and the probability defining the risk was expressed as the expectation of a cumulative return.
% % % % % % % % % % % % % % % % % % % % %

\section{CONCLUSIONS AND FUTURE WORK}
We propose a scheme using functional approximator and RL methods to solve SDM problems with risk-sensitive constraints in a dynamic scenario. %, we propose a scheme using approximator and RL methods to evaluate (estimate) 
We consider risk measures as functions of mean and variance of the return in an induced Markov process.
As shown in Section~\ref{secRisk}, most, if not all, law-invariant risk measures can be evaluated or estimated with return variance.
Considering that reward functions in practical problems are often stochastic, we implement SAT to enable the variance formula and preserve the reward sequence. 
In an inventory control problem with practical factors, an NN is trained and validated with a synthetic training dataset in the numerical experiment.

The current work can be extended in two ways.
%At least two open problems should be deliberated.
The first one has to do with the ``no free lunch'' theorem.
In our inventory control model, the more factors considered, the larger size the policy space has.
When the warehouse capacity is 3, there are 8748 deterministic policies, when it is 4 and 5, the size goes to around $ 2.4 \times 10^{7} $ and $ 1.4 \times 10^{12} $, respectively.
How to deal with a policy space of such an astronomical magnitude in a risk-sensitive case is still an unsettled problem.
The second one is how to deal with two different output data types in an NN efficiently.
In our setting, the outputs include optimal risk value (numeric) and policy (categorical). 
A crucial task is how to set activation functions, normalization, and loss function for network training.

\bibliographystyle{ieeetr}
\bibliography{bookChapter_draft.bib}

\end{document}